\documentclass[11pt]{article}

\usepackage[a4paper,margin=1in]{geometry}

\usepackage[utf8]{inputenc}
\usepackage[T1]{fontenc}
\usepackage{lmodern} %
\usepackage{microtype} %

\usepackage{setspace}
\usepackage{parskip} %

\usepackage{titlesec}
\usepackage{fancyhdr}
\usepackage{enumitem}

\usepackage{graphicx}
\usepackage{xcolor}
\usepackage{colortbl}

\usepackage{booktabs} %
\usepackage{tabularx}
\usepackage{subcaption}
\usepackage{wrapfig}
\usepackage{stfloats}
\usepackage{adjustbox}
\usepackage{hhline}

\usepackage{amsmath}
\usepackage{amsfonts} %
\usepackage{nicefrac} %

\usepackage{csquotes} %
\usepackage{hyperref}
\usepackage{xspace}
\usepackage{tcolorbox}
\usepackage{algorithm}
\usepackage{algpseudocode}
\usepackage{xparse}
\usepackage{listings}
\usepackage{minted}
\usepackage{siunitx}  

\usepackage[backend=biber,style=numeric,sorting=nyt]{biblatex}
\definecolor{sfblue}{HTML}{00A1E0} %
\definecolor{sfnavy}{HTML}{032D60} %
\definecolor{sfgray}{HTML}{706E6B} %
\definecolor{sflightblue}{HTML}{EAF5FC} %

\hypersetup{
  colorlinks=true,
  linkcolor=sfblue,
  citecolor=sfnavy,
  urlcolor=sfblue
}

\newcommand{\github}{\raisebox{-1.5pt}{\includegraphics[height=1.05em]{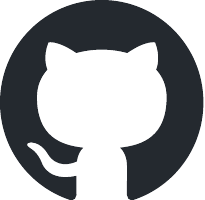}}\xspace}

\newcommand{\huggingface}{\raisebox{-1.5pt}{\includegraphics[height=1.05em]{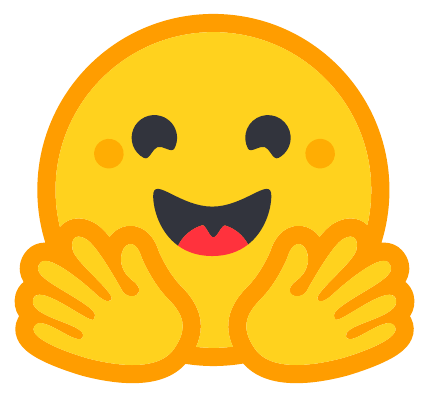}}\xspace}

\fancypagestyle{plain}{
  \fancyhf{} %
  \fancyfoot[C]{\textcolor{sfgray}{\thepage}} %
  
}

\newcommand{\legendbox}[1]{\fcolorbox{black}{#1}{\rule{0pt}{6pt}\rule{6pt}{0pt}}}

\makeatletter
\renewcommand{\maketitle}{
  \thispagestyle{plain} %
  \noindent
  \vspace*{-10pt}
  \noindent\raisebox{0pt}[0pt][0pt]{\includegraphics[height=1cm]{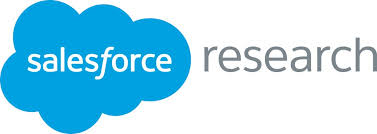}}

  \vspace{-5pt}
  \color{sfgray}\rule{\linewidth}{0.6pt}

  \vspace{10pt}
  {\Huge\bfseries\color{sfnavy} \@title \par}
  \vspace{0.5em}
  {\large \@author \par}
  \vspace{0.2em}
  \@thanks
  {\normalsize Salesforce AI Research \par}
  \vspace{1.2em}
}
\makeatother

\titleformat{\section}{\large\bfseries\color{sfnavy}}{\thesection}{1em}{}
\titleformat{\subsection}{\normalsize\bfseries\color{sfnavy}}{\thesubsection}{1em}{}

\title{\huge \textbf{E}nterprise \textbf{D}eep \textbf{R}esearch: Steerable Multi-Agent Deep Research for Enterprise Analytics}
\author{\bfseries{Akshara Prabhakar, Roshan Ram, Zixiang Chen, Silvio Savarese, Frank Wang\thanks{Work done while at Salesforce AI Research.}, Caiming Xiong, Huan Wang, Weiran Yao}}
\date{\today}
\begin{document}

\maketitle

\begin{tcolorbox}[colback=sflightblue!60,
                  colframe=sfblue,
                  boxrule=0.6pt,
                  arc=2mm,
                  left=6pt,right=6pt,top=6pt,bottom=6pt]
\textbf{Abstract. }
As information grows exponentially, enterprises face increasing pressure to transform unstructured data into coherent, actionable insights.
While autonomous agents show promise, they often struggle with domain-specific nuances, intent alignment, and enterprise integration.
We present \textbf{Enterprise Deep Research (EDR)}, a multi-agent system that integrates (1) a Master Planning Agent for adaptive query decomposition, (2) four specialized search agents (General, Academic, GitHub, LinkedIn), (3) an extensible MCP-based tool ecosystem supporting NL2SQL, file analysis, and enterprise workflows, (4) a Visualization Agent for data-driven insights, and (5) a reflection mechanism that detects knowledge gaps and updates research direction with optional human-in-the-loop steering guidance. These components enable automated report generation, real-time streaming, and seamless enterprise deployment, as validated on internal datasets.
On open-ended benchmarks including DeepResearch Bench and DeepConsult, EDR outperforms state-of-the-art agentic systems without any human steering. We release the EDR framework and benchmark trajectories to advance research on multi-agent reasoning applications.

\vspace{1em}
\begin{center}
\begin{tabular}{rll}
    \github & \textbf{\small{Code}} &
    \href{https://github.com/SalesforceAIResearch/enterprise-deep-research}{\small{\texttt{github.com/SalesforceAIResearch/enterprise-deep-research}}}\\
    \huggingface & \textbf{\small{Dataset}} & \href{https://huggingface.co/datasets/Salesforce/EDR-200}{\small{\texttt{huggingface.co/datasets/Salesforce/EDR-200}}}\\
\end{tabular}
\end{center}
\end{tcolorbox}

\color{black}

\section{Introduction}

Recent advances in LLMs have driven the emergence of general-purpose language agents capable of performing a wide range of reasoning-intensive tasks---including research assistance \cite{schmidgall2025agentlaboratoryusingllm,he-etal-2025-pasa}, retrieval-augmented generation (RAG) \cite{singh2025agentic}, coding \cite{yang2024sweagentagentcomputerinterfacesenable,yang2023intercodestandardizingbenchmarkinginteractive,}, and knowledge synthesis. At the enterprise level, such capabilities have spurred growing demand for autonomous systems that can conduct an extensive search of available resources relevant to a given prompt and produce structured, markdown-formatted reports with evidence. This has motivated a new class of \textit{deep research agents} \cite{citron2025try, perplexity2025deep, openai2025deep}---autonomous LLM-based systems that emulate human analysts by iteratively retrieving evidence, decomposing problems, and constructing higher-order insights. 

In open-ended deep research, agents must reason across unbounded corpora and evolving goals, where a “complete” answer may require synthesizing hundreds of heterogeneous documents rather than locating a single fact \cite{li2025webweaver}. The objective is to take a user-provided research prompt, perform extensive web-scale or domain-specific searches, and generate a structured report aligned with the prompt’s requirements—far beyond the scope of typical RAG systems or conversation-oriented tool-calling agents \cite{prabhakar2025apigenmt, zhang2024xlamfamilylargeaction}.  Enterprise settings amplify these challenges \cite{choubey2025benchmarkingdeepsearchheterogeneous}: information is dispersed across emails, databases, and reports; domain knowledge shifts rapidly; and goals are strategic rather than factual. Tasks like feature adoption forecasting or productivity diagnostics demand contextual reasoning, long-horizon planning, and transparent evidence attribution—areas where conventional LLM pipelines yield shallow, unverifiable outputs.

Despite rapid progress, most deep research systems remain opaque and inflexible. They operate as black boxes where users cannot inspect intermediate reasoning states, source provenance, or decision trajectories once execution begins. Although some systems permit limited plan editing before launch \cite{GeminiResearch}, they lack real-time steering—the capability that enables users to dynamically guide the agent’s reasoning trajectory--if an agent misinterprets intent, confuses entities, or diverges from task goals, it seldom recovers without costly manual restarts. Such rigidity leads to redundant API calls, inefficient exploration, and results that deviate from user intent \cite{zhang2025deep,qian2025userbenchinteractivegymenvironment,qian2025userrltraininginteractiveusercentric}. Effective research systems should instead make reasoning transparent and steerable, enabling users to intervene mid-process to correct direction, refine scope, or prioritize evidence. This interactivity narrows the search space, conserves computation, and aligns outcomes more tightly with domain constraints and user expectations.

To address these limitations, we present \textbf{E}nterprise \textbf{D}eep \textbf{R}esearch \textbf{(EDR)}—a transparent, steerable multi-agent framework for adaptive, interpretable, and user-aligned research. Drawing on Anthropic’s notion of \textit{thinking in context} \cite{anthropic2025context}, EDR formalizes \textit{steerable context engineering}, enabling humans to modify agent context dynamically during execution. Steering occurs at the context curation layer—directly influencing what information enters the agent’s attention at each decision point—rather than through pre-specified constraints or post-hoc corrections \cite{NVIDIA-AI-Q}. By exposing the agent’s internal planning state (via \texttt{todo.md}) and translating natural language steering inputs into context modifications—such as task additions, cancellations, or reprioritizations—EDR empowers users to act as \textit{context curators} instead of passive observers. The framework ensures that reasoning remains \textit{visible} (through explicit task and provenance tracking), \textit{modifiable} (via real-time steering), and \textit{traceable} (with full evidence transparency). This design establishes a dynamic human–AI collaboration loop that maintains contextual grounding over long horizons, mitigates lost-in-the-middle issues \cite{liu2023lost}, and ultimately delivers faithful and scalable research for enterprise applications.

\begin{figure}[ht]
\centering
\includegraphics[width=\linewidth]{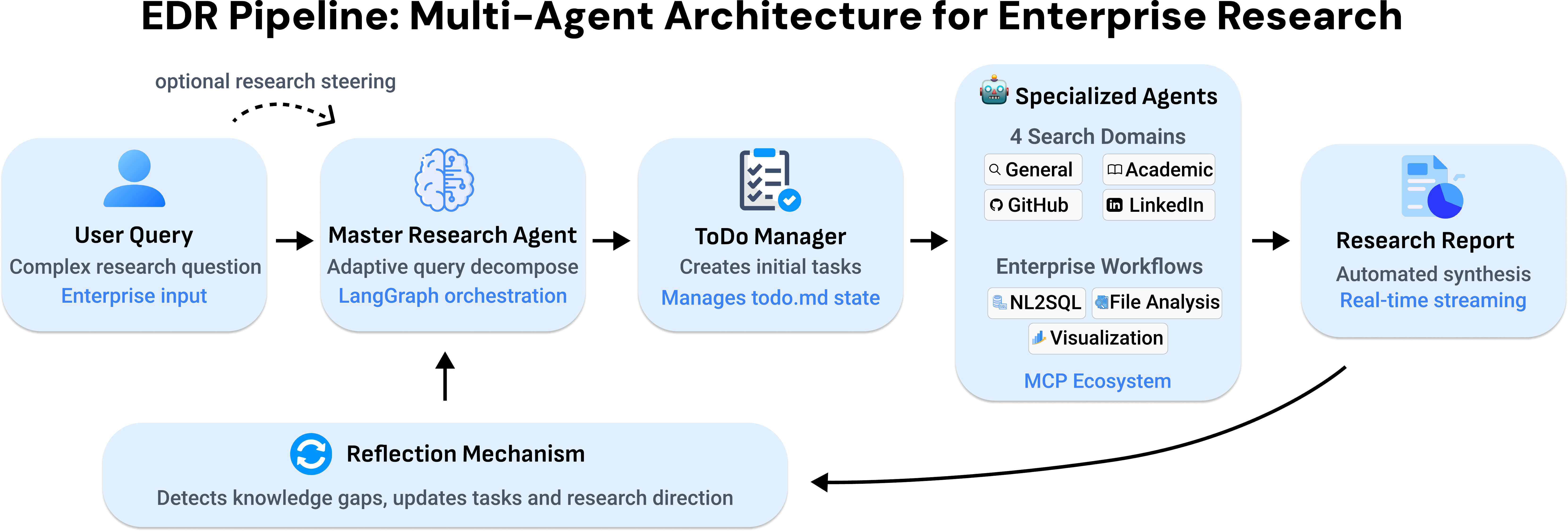}
    \caption{\small{Overview of \textbf{E}nterprise \textbf{D}eep \textbf{R}esearch framework---orchestrated system combining planning, specialized search, extensible 
enterprise tools, visualization, and reflection with optional human steering.}}
\label{fig:pipeline}
\end{figure}

Our key contributions are threefold:
\begin{itemize}[leftmargin=1.5em, noitemsep, topsep=0pt]
\item we introduce EDR (\autoref{fig:pipeline}), a modular, extensible, multi-agent architecture that is fully configurable for optimal research execution for enterprise;
\item we provide users fine-grain control over the research process via a todo-driven steering framework that enables human-in-the-loop guidance during research execution, not just at the planning stage;
\item we open-source EDR-200, a dataset comprising EDR's complete research trajectories from benchmark evaluations to promote future research.
\end{itemize}

\section{System Architecture}

Enterprise Deep Research (EDR) adopts a modular, multi-agent architecture engineered for scalability, robustness, and adaptive orchestration of complex research workflows (\autoref{fig:pipeline}). The framework integrates a central planning agent, structured task management, domain-specialized retrieval tools, and extensible enterprise connectors, enabling iterative, human-in-the-loop research at scale. Prompts used across all stages are detailed in \autoref{sec:prompts}.

\subsection{Master Research Agent}

The Master Research Agent functions as the central orchestrator, decomposing high-level research objectives into discrete tasks and coordinating multi-agent execution. Leveraging function calling and context-aware prompt engineering, the agent performs adaptive query decomposition, intent classification, and entity extraction. Each task is annotated with priority, recommended tools, metadata, and execution dependencies.  

Decomposition integrates three contextual signals: (1) the user query, (2) knowledge gaps identified from prior iterations, and (3) steering directives issued by the Research Todo Manager. 
Dynamic replanning permits adjustments based on intermediate results, agent efficacy, or updated directives.  
Quality control is embedded within the Master Agent. Mechanisms include semantic consistency validation, cross-agent result comparison, automated fact-checking, and confidence scoring. Multi-agent outputs are aggregated into coherent interim summaries, resolving conflicts, normalizing citations, and compressing context while preserving provenance.

\paragraph{Adaptive Decomposition.}
The agent employs LLM function calling to classify user queries as simple or complex. Simple queries (e.g., ``What is GPT-4?'') yield single targeted searches. Complex queries (e.g., ``AI's impact on healthcare, education, and employment'') decompose into parallel tasks with independent search strategies. In subsequent loops, the LLM receives all pending tasks, preventing redundant work through explicit duplicate-awareness.

\paragraph{Temporal and Knowledge Grounding.}
Prompts embed temporal markers (\texttt{CURRENT\_DATE}, \texttt{CURRENT\_YEAR}) to ground recency-sensitive queries. User-uploaded documents receive authoritative status, guiding the agent to generate complementary—not redundant—searches that validate or extend internal knowledge with external evidence.

\subsection{Research Todo Manager}

The Research Todo Manager provides a structured, human-interpretable task management layer. Tasks are stateful objects with unique IDs, natural language descriptions, priority scores (5--10), lifecycle status (\textit{pending}, \textit{in-progress}, \textit{completed}, \textit{canceled}), and provenance tags (\texttt{initial\_query}, \texttt{knowledge\_gap}, \texttt{steering\_message}).
The \texttt{todo.md} file functions as a persistent, shared representation between agents and human users, facilitating transparency and collaborative context management; functioning as both agent execution plan and user progress tracker.  
Where Manus \cite{manus2025context} employs todos solely for agent planning, EDR extends this approach by exposing todos to human oversight, combining structured task management with persistent note-taking. This hybrid design allows both agent and human to collaboratively maintain context and steer research direction.  

Priority-based scheduling ensures high-impact research items are addressed first, mitigating context dilution during long-horizon LLM planning. Users may issue natural language directives (e.g., ``focus on peer-reviewed sources,'' ``prioritize recent publications''), which are translated into task adjustments, exclusions, or priority modifications. Versioning enables efficient frontend synchronization, maintaining a consistent view of system progress without continuous state streaming.  

\subsection{Search API and Tools}

\paragraph{General Web Search.}
Retrieves broad web content, including news, reports, and general knowledge. Implements top-k result retrieval with full content extraction, semantic deduplication, and relevance scoring. Integrates the Tavily API to provide high-throughput, structured access to web content.  

\paragraph{Academic Search.}
Targets scholarly publications and peer-reviewed content. Supports higher result limits, fuzzy deduplication for title variations, and optional temporal weighting to emphasize recent research. Integrates multiple academic repositories, including arXiv, for comprehensive coverage and ensures structured metadata extraction for downstream processing.  

\paragraph{GitHub Search.}
Focuses on code repositories, technical implementations, and software documentation. Employs repository-level deduplication to prevent redundancy and prioritizes file-level URLs for maximum utility. Direct API integration with GitHub ensures reliability and high throughput, while results include structured metadata for seamless integration with domain agents.  

\paragraph{LinkedIn Search.}
Extracts professional profiles, company information, and domain expertise. Optimized for top-K retrieval, selective raw content retention, and strict domain restriction to linkedin.com, enabling precise enterprise-focused research queries. Metadata extraction supports downstream evaluation of expertise and relevance.  

\subsection{Domain-Specific and MCP Tools}

\paragraph{File Analysis.}
Processes structured and unstructured files (db, sqlite, pdf, docx, txt, csv, xlsx, images) using format-specific parsers and LLM-powered content summarization. Includes metadata extraction, layout preservation, and semantic content analysis to integrate uploaded knowledge into the research context.  

\paragraph{NL2SQL Agent.}
Translates natural language queries into SQL statements for structured database interrogation. Incorporates schema awareness, query decomposition, multi-layered validation (syntax, semantics, performance, security), and result interpretation. Optimized for enterprise data warehouses, it supports context-sensitive filtering and automated reasoning over relational schemas.  

\paragraph{Visualization Agent.}
Generates visual representations from quantitative findings. Chart types are selected adaptively based on data characteristics, including bar, line, scatter, heatmap, and pie charts. Visualizations are rendered in sandboxed execution environments and support interactive exploration, export to multiple formats (pdf, docx, html, png, svg), and structured report integration.  

\paragraph{Enterprise Connectors via MCP.}
EDR supports extensible integration through the Model Context Protocol (MCP), enabling connection to custom enterprise systems, remote computation services, and additional domain tools. MCP supports HTTP and stdio transport for universal tool interoperability, reducing integration overhead for new agents. Supported examples include code search, database querying, and organization-specific computational tools.  

\subsection{Integration and Workflow Coordination}

EDR tightly couples task decomposition, search execution, and domain tool utilization within iterative research loops. Outputs from all agents are consolidated into a unified running summary and task reflection mechanism, continuously informing the Research Todo Manager and Master Agent. This design preserves context coherence, supports cumulative knowledge integration, and provides a seamless transition into the Research Flow Mechanism, enabling systematic report generation from complex, multi-domain queries.

\section{Research Flow Mechanism}

EDR translates high-level user queries into comprehensive analytical reports through a structured, iterative workflow that integrates human-in-the-loop steering with multi-agent execution. The API specifications and frontend implementation is detailed in \autoref{sec:system-architecture}.

\subsection{User Query Ingestion and Todo Initialization}
\begin{figure}[t]
\centering
\begin{minipage}[c]{0.55\textwidth}
  \flushright
  \includegraphics[height=0.95\linewidth]{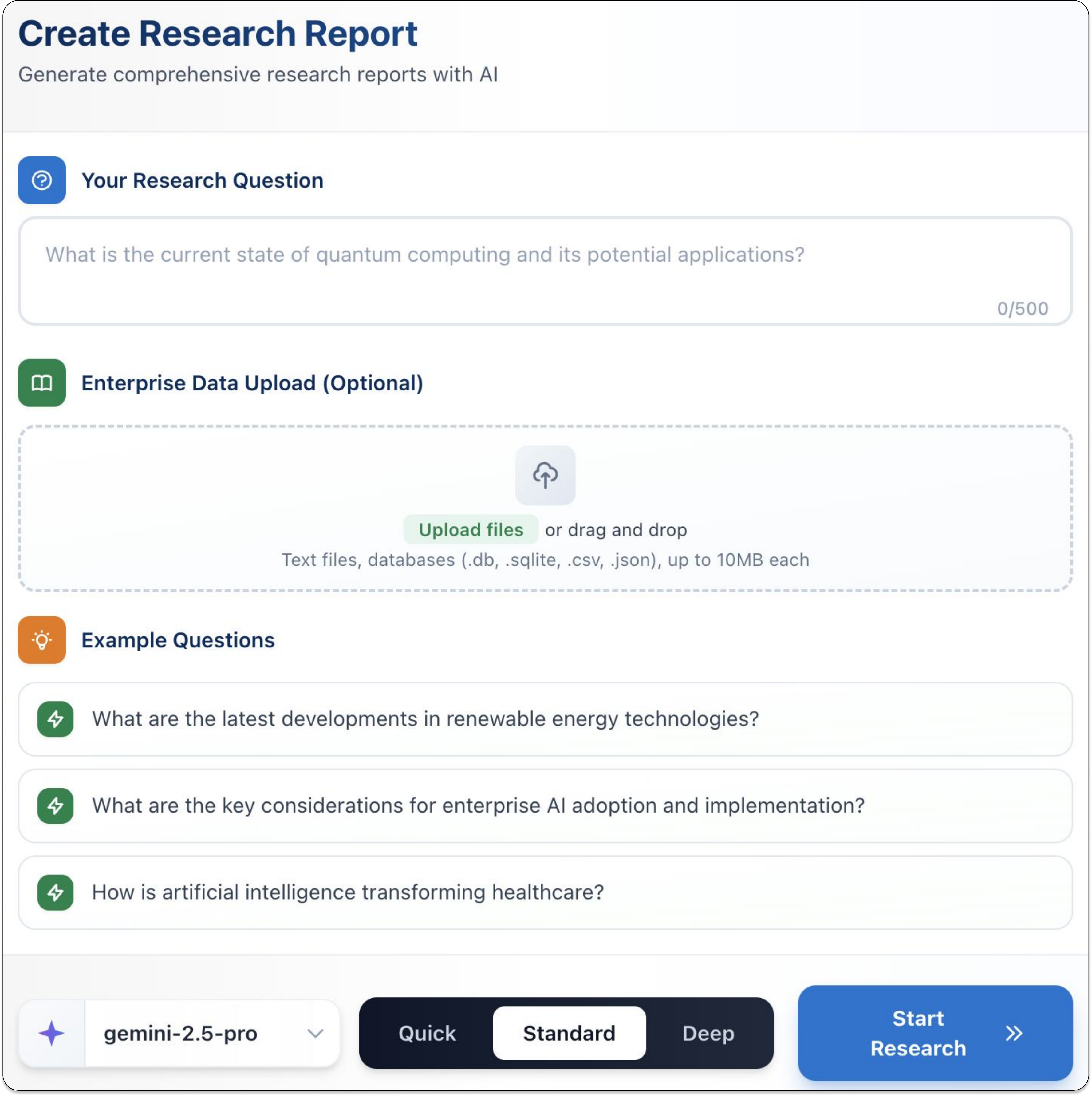}
\end{minipage}
\hfill
\begin{minipage}[c]{0.36\textwidth}
  \caption{\small Screenshot of EDR's home screen UI. This is the initial screen where users can submit their research query and add optional data files (e.g., images, sheets, databases). The research offers 3 run modes: \textit{quick} (for a high-level investigation), \textit{standard} (the standard deep research mode), and \textit{deep} (for a max-effort report) and allows the user to choose their underlying base LLM.}
  \label{fig:edr_ui_left}
\end{minipage}
\end{figure}

Upon submission, the user query is received via the frontend interface (\autoref{fig:edr_ui_left}) and a new session is spawned. The \texttt{ResearchTodoManager} constructs an initial lightweight 3--5 task plan for the user query (Appendix \autoref{fig:initial_task_prompt}), providing immediate UI feedback. Each task is annotated with a unique identifier, a priority score (ranging 5--10, either LLM-suggested or calculated as $5 + (N - i)$ where $N$ is the total subtask count and $i$ is the task index), and provenance metadata tracking its source.  Tasks originating from the initial user query are labeled \texttt{initial\_query}, while those generated in subsequent loops to address knowledge gaps are tagged \texttt{knowledge\_gap}. Tasks created in response to user steering messages are marked \texttt{steering}, with steering-derived tasks receiving the highest priority (10), followed by original query tasks (9), and knowledge-gap tasks (7). The \texttt{ResearchTodoManager} maintains these tasks in a persistent \texttt{todo.md} representation as a named variable in the code execution state. As the research progresses, EDR updates it and checks off completed items.

\subsection{Task-Query Transformation}
At the beginning of each research iteration, the \texttt{MasterResearchAgent} constructs an integrated prompt (Appendix \autoref{fig:primary_query_generation_prompt}) embedding the original research objective, the progressively refined summary of prior findings, the set of unresolved information needs, and all active steering constraints derived from user interactions. Guided by this comprehensive context, the LLM performs adaptive query decomposition, generating 3--7 new tasks and corresponding search queries, with task source transitioning from \texttt{initial\_query} (Loop 0) to \texttt{knowledge\_gap} (Loop 1+). The agent's decomposition prompt includes existing pending tasks, instructing the LLM to avoid redundant work—though fuzzy matching provides a safety net for any remaining overlaps. Each proposed search query includes metadata specifying the recommended retrieval domain (e.g., academic\_search for scholarly content, github\_search for code, general\_search for broad topics, nl2sql for database queries) and a descriptive aspect indicating the query's focus area. 

Before dispatching these search queries, three layers of quality control are enforced. First, semantic deduplication prevents redundant searches by fuzzy string matching with prefix normalization, merging duplicates and updating their priority if higher. Second, constraint enforcement ensures compliance with steering instructions: for instance, excluding terms marked by the user or elevating those aligned with focus directives. Finally, priority adjustment dynamically reorders the execution sequence so that high-impact and user-aligned queries receive preferential processing. Tasks associated with selected queries are transitioned to the \textit{in-progress} state, ensuring consistent synchronization between planning intent and execution progress across research loops. Tasks transition through a formal lifecycle: \textit{pending} (awaiting execution) $\rightarrow$ \textit{in-progress} (actively querying) $\rightarrow$ \textit{completed} (successfully resolved) or \textit{canceled} (rendered obsolete). Queries are dispatched in parallel to specialized agents, including general search, academic literature, code repositories, and domain tools (NL2SQL, MCP tools) as appropriate, ensuring domain-specific execution efficiency. Agents perform filtering, deduplication, and relevance scoring before returning results for aggregation.

\subsection{Result Aggregation and Incremental Synthesis}
\label{sec:synthesis}

Returned results from domain-specific agents undergo three-stage processing:

\paragraph{Stage 1: Inter-Agent Deduplication.}
Results are consolidated through semantic similarity comparison, identifying overlapping content across multiple search tools. Citation normalization ensures consistent URL and title formatting, with priority given to the highest-quality representation of each unique source.

\paragraph{Stage 2: LLM-Driven Synthesis.}
In this stage, the LLM merges newly gathered research content into the existing running summary using a dedicated synthesis process. It takes four inputs: (1) the previous iteration's running summary, (2) newly fetched web research results, (3) knowledge gaps from reflection, and (4) user-uploaded knowledge (if present). The LLM performs context compression, extracting key insights while preserving citation links and metadata. This prevents exponential context growth—rather than accumulating all raw search outputs, the system maintains a progressively refined knowledge representation across iterations.

\paragraph{Stage 3: Source Citation Management.}
Extracted sources are tracked in a deduplicated dictionary, maintaining URL-to-metadata mappings for downstream report generation. Unused sources are logged for transparency but excluded from final citations.

The synthesis step ensures that each iteration builds upon prior findings without losing coherence or exceeding context limits, enabling sessions with 10+ iterations and hundreds of sources.

\subsection{Steering Integration}
EDR implements a queue-based, race-condition-safe steering mechanism that enables real-time user guidance without interrupting ongoing execution. User messages are queued during research execution and, if multiple messages accumulate, they are summarized to extract core directives (e.g., ``focus on peer-reviewed sources'' and ``prioritize recent papers'' $\rightarrow$   ``emphasize recent peer-reviewed literature''). Messages are processed atomically between iterations during the reflection phase (\autoref{sec:reflection}), preventing interference with active queries. To prevent data loss, the system employs snapshot-based merging: steering messages that arrive \emph{during} reflection are automatically preserved and appended to the post-reflection queue. This ensures user input is never lost while maintaining deterministic LLM reasoning over a stable message set. The extracted directives update the \texttt{ResearchTodoManager}; incorporate steering constraints as priority boosts, exclusion filters, or focus directives, ensuring alignment with user intent across subsequent query generation and execution cycles.

\subsection{Reflection and Todo Update}
\label{sec:reflection}
The reflection mechanism serves as the core process in EDR. After each iteration, the system evaluates the aggregated results against the current todo plan and accumulated knowledge, identifying:

\begin{itemize}[leftmargin=1.5em, noitemsep, topsep=0pt]
    \item \textbf{Knowledge Gaps:} Missing concepts, unexplored domains, or insufficient evidence relative to the original user query.  
    \item \textbf{Task Misalignment:} Tasks that are no longer relevant due to new findings or user steering directives.  
    \item \textbf{Quality Inconsistencies:} Contradictory or low-confidence information returned by agents.  
\end{itemize}

Based on this analysis, the \texttt{ResearchTodoManager} updates the todo plan (Appendix \autoref{fig:reflection_prompt}):
\begin{itemize}[leftmargin=1.5em, noitemsep, topsep=0pt]
    \item \textbf{Generate New Tasks:} New tasks are generated to address knowledge gaps, with priority scores reflecting their importance.  
    \item \textbf{Update Task Status:} Misaligned or resolved tasks are \textit{canceled} or marked as \textit{completed}.  
    \item \textbf{Clear Queue:} Specifying indices of user steering messages fully addressed through task creation/cancellation. Unaddressed messages remain queued for subsequent loops. 
\end{itemize}

\begin{figure}[h]
\centering
\includegraphics[width=\linewidth]{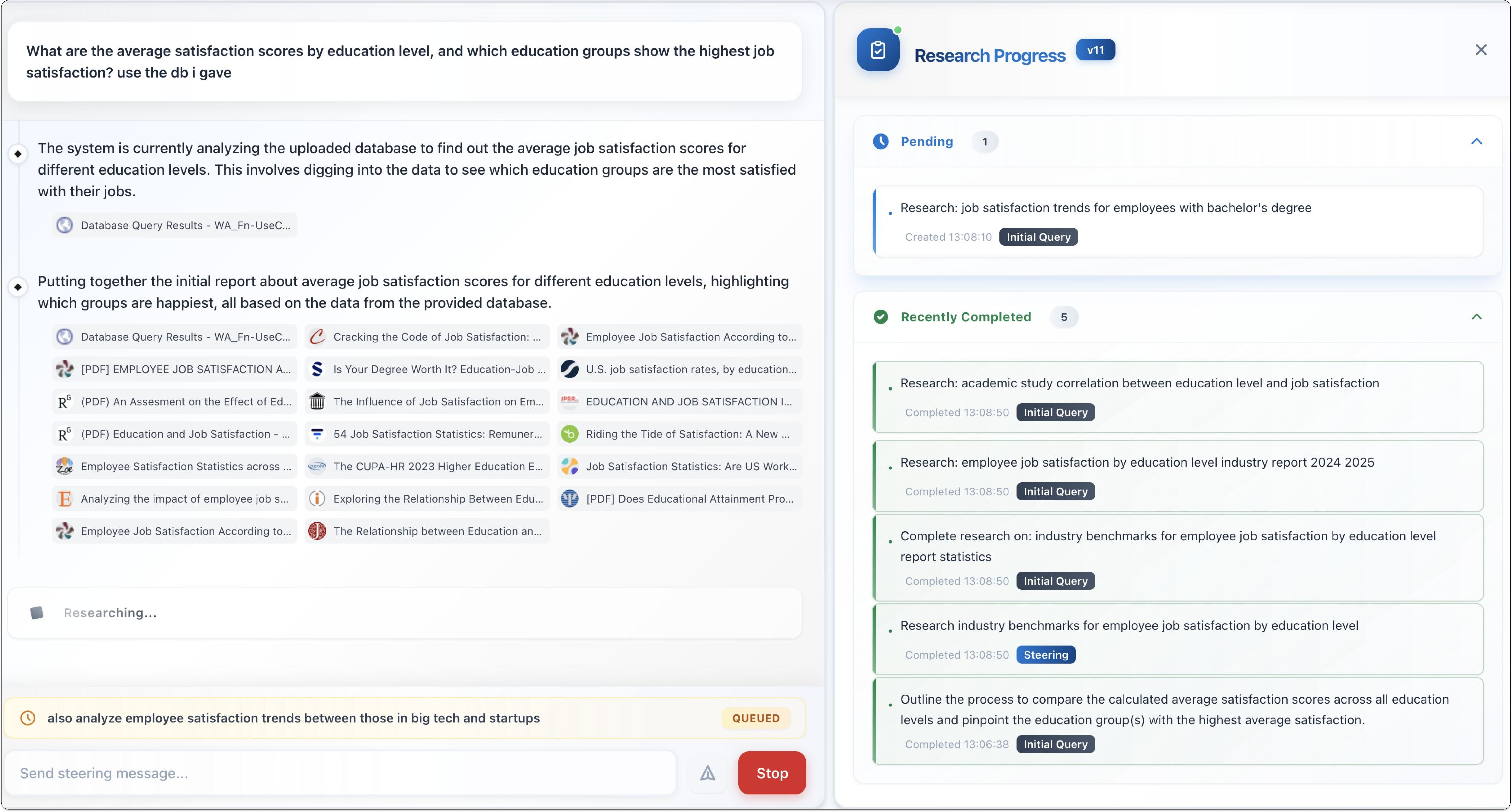}
\caption{\small{Screenshot of EDR's research execution screen. Left pane shows the progress with a summary of the current step, while the progress pane shows the todo status. Each task is annotated with its status and provenance with timestamps for full traceability. The received steering message is queued until the reflection phase.}}
\label{fig:edr_ui_right}
\vspace{-4mm}
\end{figure}

The \texttt{ResearchTodoManager} increments a version counter on every state modification, triggering frontend updates only when changes occur. This version-based polling provides real-time visibility into task status and provenance without continuous state streaming (\autoref{fig:edr_ui_right}). Reflection is iterative and cumulative, ensuring that subsequent research loops increasingly focus on unaddressed knowledge gaps while maintaining continuity with prior findings.

\subsection{Iterative Refinement and Loop Continuation}
The cycle of query planning, agent execution, steering integration, result aggregation, and reflection repeats iteratively. Each loop incorporates feedback from previous iterations and user steering messages, gradually converging toward comprehensive coverage of the research question. Termination occurs when knowledge gaps are resolved, maximum loop limits are reached, or the system determines sufficient report completeness.

\subsection{Final Report Generation and Validation}
Upon completion, the system synthesizes the running summary, aggregated sources, code snippets, and steering history into a structured report. Quality assurance checks validate citation completeness, structural coherence, query coverage, and adherence to user directives. The final document, delivered via the frontend interface, provides an interactive and detailed account of the research process, preserving transparency, reproducibility, and user-aligned insight generation.

\section{Evaluations}

\subsection{Experimental Setup}

For all evaluations, we use \texttt{gemini-2.5-pro} as the underlying base model, as it showed the strongest performance in our preliminary tests. We set the max research loops to 5 for DeepResearch Bench, 10 for DeepConsult, and 2 for ResearchQA. Real-time steering is kept disabled as the research query is fixed for these evaluation settings, with no human intervention.

\paragraph{Baselines.}
We test against popular proprietary deep research assistants OpenAI DeepResearch \texttt{openai-deepresearch} \cite{openai2025deep}, Gemini-2.5-pro-deep-research \cite{GeminiResearch}, Claude-research \cite{claude}, Perplexity Deep Research \cite{perplexity2025deep}, doubao-research \cite{doubao2025deepresearch}, kimi-research \cite{kimi2025deepresearch}
and open source systems like langchain-open-deep-research \cite{langchain-odr}, WebShaper \cite{tao2025webshaper}, and the more recent WebWeaver \cite{li2025webweaver}.

\paragraph{Datasets.} We evaluate on 3 popular open-ended deep research benchmarks:

\textbf{(1) DeepResearch Bench \cite{du2025deepresearch}}, comprising 100 PhD-level complex research tasks meticulously formulated by domain experts across 22 distinct fields, such as Science \& Technology, Finance \& Business, Software Engineering, and Art \& Design.

\textbf{(2) DeepConsult \cite{deepconsult}}, a specialized collection of business and consulting-focused prompts designed for in-depth research, encompassing a broad range of topics such as marketing strategy, financial analysis, emerging technology trends, and business planning.

\textbf{(3) ResearchQA \cite{yifei2025researchqa}}, a large-scale, multi-field benchmark comprising 3750 scientific test questions curated by PhD scholars mined from academic survey papers spanning 7 broad domains. Each query is annotated with one or more rubrics which are scored to indicate the extent to which a response satisfies the criterion, enabling detailed, fine-grained analysis across domains and question types.

\subsection{Results}

\textbf{DeepResearch Bench Results.  } It employs two evaluation metrics \cite{du2025deepresearch}: RACE (Report Quality) measures the quality of generated reports against references across four dimensions—Comprehensiveness, Insight, Instruction-Following, and Readability—aggregated into an overall weighted score represented by Over. Citation Accuracy (CitAcc.) evaluates retrieval accuracy and reliability. Following the benchmark setup, the judge model for the comparison is Gemini-2.5-pro. EDR obtains exceptional performance on the DeepResearch Bench leaderboard. As shown in \autoref{tab:drb}, EDR outperforms all proprietary and most open-source agentic systems with an overall score of 49.86. Specifically, on the Instruction Following and Readability criteria, EDR obtains particularly high scores. In terms of cost, we see that EDR consumes 4x lesser tokens than \texttt{langchain-open-deep-research}.

\begin{table}[htbp]
\centering
\scriptsize
\setlength{\tabcolsep}{3pt} %
\begin{tabularx}{\textwidth}{
    l
    S[table-format=2.2]
    S[table-format=2.2]
    S[table-format=2.2]
    S[table-format=2.2]
    S[table-format=2.2]
    S[table-format=3.2]
    S[table-format=9.0]
    S[table-format=4.2]
}
\toprule
\textbf{Agentic System} & 
\textbf{Over.} & 
\textbf{Comp.} & 
\textbf{Ins.} & 
\textbf{Inst.} & 
\textbf{Read.} & 
\textbf{CitAcc.} & 
\textbf{Tokens} & 
\textbf{Cost (\$)} \\
\midrule
gemini-2.5-pro-preview-05-06              & 31.90 & 31.75 & 24.61 & 40.24 & 32.76 & \text{-} & \text{-} & \text{-} \\
WebShaper (32B)                           & 34.93 & 31.58 & 26.17 & 44.81 & 40.38 & \text{-} & \text{-} & \text{-} \\
langchain-open-deep-research              & 43.44 & 42.97 & 39.17 & 48.09 & 45.22 & \text{-} & \text{207,005,549} & 87.83 \\
doubao-research                           & 44.34 & 44.84 & 40.56 & 47.95 & 44.69 & 52.86 & \text{-} & \text{-} \\
kimi-research                             & 44.64 & 44.96 & 41.97 & 47.14 & 45.59 & \text{-} & \text{-} & \text{-} \\
Claude-research                           & 45.00 & 45.34 & 42.79 & 47.58 & 44.66 & \text{-} & \text{-} & \text{-} \\
openai-deepresearch                       & 46.45 & 46.46 & 43.73 & 49.39 & 47.22 & 75.01 & \text{-} & \text{-} \\
\rowcolor{gray!15} WebWeaver (qwen3-30b-a3b-instruct-2507)  & 46.77 & 45.15 & 45.78 & 49.21 & 47.34 & 25.00 & \text{-} & \text{-} \\
\rowcolor{gray!15} WebWeaver (gpt-oss-120b)                  & 48.11 & 48.03 & 47.20 & 48.94 & 48.11 & 66.14 & \text{-} & \text{-} \\
Gemini-2.5-pro-deepresearch               & 49.71 & 49.51 & 49.45 & 50.12 & \underline{50.00} & 78.30 & \text{-} & \text{-} \\
\rowcolor{gray!15} WebWeaver (Claude-sonnet-4-20250514)      & \underline{50.58} & \textbf{51.45} & 50.02 & \textbf{50.81} & 49.79 & \textbf{93.37} & \text{71,922,021} & \text{-} \\
\rowcolor{gray!15} WebWeaver (qwen3-235b-a22b-instruct-2507) & \textbf{50.62} & \underline{51.29} & \underline{51.00} & 49.98 & 48.89 & \underline{78.25} & \text{-} & \text{-} \\
\midrule
\textbf{Enterprise Deep Research} & \textbf{50.62} & \text{49.70} & \textbf{51.24} & \underline{50.52} & \textbf{50.61} & 72.50 & \textbf{\text{53,926,192}} & 117.48 \\
\bottomrule
\end{tabularx}
\caption{\small Performance on DeepResearch Bench. Best number shown in \textbf{bold}, second best is \underline{underlined}. 
\legendbox{gray!15} indicates the system is not accessible to test.}
\label{tab:drb}
\end{table}

\textbf{DeepConsult Results.   } \autoref{tab:dc} presents the performance of agentic systems on DeepConsult. Evaluation is conducted via pairwise comparison against the OpenAI DeepSearch baseline, using win, tie, and loss rates as primary metrics, along with an average quality score. The judge model used is \texttt{gpt-4.1-2025-04-14}.  We observe that EDR achieves the highest win rate of 71.57\% and a superior average score of 6.82, which is higher than other open-source systems. Additionally, the lose rate is much lower at 9\% indicating that most reports are significantly superior or at par with \texttt{openai-deepresearch}. Overall, one run with EDR takes 48,850,862 tokens amounting to a total cost of \$105.51.

\begin{table}[h]
\centering
\scriptsize
\setlength{\tabcolsep}{5.9pt}
\begin{tabularx}{\textwidth}{
    l
    S[table-format=2.2]
    S[table-format=2.2]
    S[table-format=3.2]
    S[table-format=2.2]
}
\toprule
\textbf{Agentic System} & 
\textbf{Win Rate (\%)} & 
\textbf{Tie Rate (\%)} & 
\textbf{Lose Rate (\%)} & 
\textbf{Avg. Score} \\
\midrule
WebShaper (32B) & 3.25 & 3.75 & 93.00 & 1.63 \\

\rowcolor{gray!15}
WebWeaver (qwen3-30b-a3b-instruct-2507) &
\multicolumn{1}{S[table-format=2.2]}{28.65} &
\multicolumn{1}{S[table-format=2.2]}{34.90} &
\multicolumn{1}{S[table-format=3.2]}{36.46} &
\multicolumn{1}{S[table-format=2.2]}{4.57} \\

Claude-research & 25.00 & 38.89 & 36.11 & 4.60 \\
doubao-research & 29.95 & 40.35 & 29.70 & 5.42 \\
Perplexity Deep Research & 32.00 & {-} & {-} & {-} \\
openai-deepresearch & 0.00 & 100.00 & 0.00 & 5.00 \\

\rowcolor{gray!15}
WebWeaver (qwen3-235b-a22b-instruct-2507) &
\multicolumn{1}{S[table-format=2.2]}{54.74} &
\multicolumn{1}{S[table-format=2.2]}{28.61} &
\multicolumn{1}{S[table-format=3.2]}{16.67} &
\multicolumn{1}{S[table-format=2.2]}{6.47} \\

Gemini-2.5-pro-deepresearch & 61.27 & 31.13 & \textbf{7.60} & 6.70 \\

\rowcolor{gray!15}
WebWeaver (gpt-oss-120b) &
\multicolumn{1}{S[table-format=2.2]}{65.31} &
\multicolumn{1}{S[table-format=2.2]}{11.22} &
\multicolumn{1}{S[table-format=3.2]}{23.47} &
\multicolumn{1}{S[table-format=2.2]}{6.64} \\

\rowcolor{gray!15}
WebWeaver (Claude-sonnet-4-20250514) &
\multicolumn{1}{S[table-format=2.2]}{66.86} &
\multicolumn{1}{S[table-format=2.2]}{10.47} &
\multicolumn{1}{S[table-format=3.2]}{22.67} &
\multicolumn{1}{S[table-format=2.2]}{\textbf{6.96}} \\

\midrule
\textbf{Enterprise Deep Research} & 
\textbf{71.57} & 19.12 & \underline{9.31} & \underline{6.82} \\
\bottomrule
\end{tabularx}
\caption{\small Performance on DeepConsult. Best number shown in \textbf{bold}, second best is \underline{underlined}. 
\legendbox{gray!15} indicates systems that are not accessible to test.}
\label{tab:dc}
\end{table}

\textbf{ResearchQA Results.   } This benchmark evaluates research responses across six distinct rubric categories -- Citation, Impact, Comparison, Example, Limitation, or Other. Queries can be tagged with multiple types, capturing intersections like Comparison + Impact or Example + Impact, which allows assessment of both individual capabilities and integrated reasoning across multiple research competencies. From \autoref{tab:rqa}, we see that Perplexity Deep Research (Sonar) achieves the highest overall coverage at 75.3\%, while EDR attains a competitive 68.5\%. Analysis by rubric type reveals EDR's strong performance on General, Impact, and Comparison items, but severe weaknesses in citations (85\% failures), example generation, and multi-criteria rubrics like Comparison+Example+Impact. Domain-specific patterns emerge, with Life \& Earth Sciences and Business \& Economics performing best, whereas Humanities \& Arts struggles with examples. Score distributions are strongly bimodal, indicating that responses either fully satisfy a rubric item or fail completely. These findings highlight critical areas for improvement, specifically citation handling and example generation to enhance EDR’s reliability for scientific research tasks.

\begin{table}[htbp]
\centering
\scriptsize
\setlength{\tabcolsep}{3pt}
\begin{tabularx}{\textwidth}{
    l
    S[table-format=3.1]
    S[table-format=2.2]
    *{7}{S[table-format=2.2]}
}
\toprule
\textbf{Agentic System} &
\textbf{Avg Length} &
\textbf{All} &
\textbf{Health.} &
\textbf{Life.} &
\textbf{Engg.} &
\textbf{PhysSci.} &
\textbf{SocSci.} &
\textbf{Human.} &
\textbf{Econ.} \\
\midrule
sonar & 242.2 & 58.61 & 56.61 & 60.55 & 59.43 & 61.62 & 59.97 & 57.48 & 62.80 \\
openscholar-8b+feedback & 788.8 & 58.72 & 57.77 & 59.96 & 58.62 & 61.48 & 57.27 & 57.29 & 62.48 \\
sonar-reasoning & 280.5 & 64.33 & 62.73 & 66.00 & 65.19 & 68.11 & 62.68 & 61.76 & 67.49 \\
gpt-4o-search-preview & 255.0 & 65.98 & 65.52 & 68.21 & 65.01 & 66.60 & 66.07 & 62.62 & 65.63 \\
gemini-2.5-pro+grounding & 278.5 & 68.51 & 67.38 & 70.02 & 68.76 & 70.99 & 68.09 & 65.98 & 71.21 \\
claude-4-sonnet+ws & 327.8 & 69.18 & 69.54 & 70.49 & 67.59 & 70.28 & 68.14 & 64.70 & 67.13 \\
o4-mini-deep-research & 271.6 & 72.69 & 74.02 & 73.58 & 70.57 & 74.04 & 73.25 & 68.99 & 74.54 \\
sonar-deep-research & 267.3 & \textbf{75.29} & \textbf{75.01} & \textbf{76.31} & \textbf{74.48} & \textbf{76.77} & \textbf{75.34} & \textbf{72.47} & \textbf{78.01} \\
\midrule
\textbf{Enterprise Deep Research}$^*$ & \textbf{220.1} & 68.52 & 67.98 & 69.70 & 68.61 & 69.80 & 68.41 & 65.11 & 70.35 \\
\bottomrule
\end{tabularx}
\caption{\small  Coverage (\%) of various agentic LLM systems across seven research domains in ResearchQA. $^*$We run EDR for a max of 2 research loops, to limit costs for running across 3K test  queries.}
\label{tab:rqa}
\end{table}

\textbf{Enterprise Usecase.   } We also conduct evaluations on internal enterprise use cases encompassing both open-ended research and research over complex internal proprietary databases. EDR achieves over 95\% accuracy in SQL generation and execution, 99.9\% uptime, while maintaining reliability and scalability across diverse workloads. User studies report a 98\% task completion rate, a 4.8/5 satisfaction score, and a 50\% reduction in time-to-insight for complex analytical tasks.

\subsection{Trajectory Collection}
We collect 201 complete agentic trajectories generated by EDR—99 on DeepResearch Bench and 102 on DeepConsult. Unlike prior benchmarks that capture only final outputs, these trajectories expose the full reasoning process—search, reflection, and synthesis—enabling fine-grained analysis of planning and decision dynamics. It aids in studying long-horizon agentic behavior and developing training and evaluation methods for more efficient research agents. \autoref{tab:data_stats} summarizes key statistics. Report synthesis peaks at iterations 4–5 with a +1,785-word gain (3× average), marking the most productive growth phase following sufficient information accumulation. Reflection analysis across 1,422 instances reveals recurring knowledge gaps in market (27.1\%), comparative (18.4\%), and cost (14.2\%) analyses. Source usage remains stable at  $\approx$ 14 per iteration, reflecting EDR’s sustained diversity and coherence throughout extended research workflows.

\begin{table}[h!]
\centering
\scriptsize
\begin{tabular}{@{}lr@{}}
\toprule
\textbf{Metric} & \textbf{Value} \\ \midrule
Avg. Iterations per Trajectory & 7.19 \\
Avg. Tool Calls per Trajectory & 49.88 \\
Avg. Tool Calls per Iteration & 6.93 \\
Avg. Searches per Trajectory & 28.30  \\
Avg. Report Length (words) & 6,523 \\
Avg. Report Growth per Iteration (words) & 600 \\
\bottomrule
\end{tabular}
\caption{\small Statistics of EDR-200 trajectory dataset.}
\label{tab:data_stats}
\end{table}

\section{Related Works}
\paragraph{AI-driven Research Systems.}

AI-driven research has advanced from short-answer retrieval pipelines to long-horizon systems that emulate human analysts through iterative reasoning and evidence synthesis. Proprietary systems such as OpenAI Deep Research, Gemini Deep Research \parencite{citron2025try}, and Claude Research \parencite{claude} achieve expert-level performance on open-domain tasks like fact-checking and report writing but remain closed-source and non-reproducible. Open efforts initially targeted benchmark-oriented QA and retrieval-augmented generation \parencite{DBLP:journals/corr/abs-2507-02592,DBLP:journals/corr/abs-2507-15061,agentfounder2025,qiao2025webresearcher}, later extending to long-form report generation via OpenDeepResearch \cite{OpenDeepResearch} and GPT Researcher \parencite{GPTResearch}. However, these systems typically follow rigid draft-then-retrieve pipelines that fix outlines before evidence collection, leading to incoherence and hallucination \parencite{ttd}. Recent frameworks like WebWeaver \parencite{li2025webweaver} and NVIDIA-AIQ \parencite{NVIDIA-AI-Q} move toward adaptive, strategy-aware research but still rely on static control and public web data. A key gap remains: current systems cannot conduct enterprise-scale deep research over proprietary, heterogeneous sources while remaining interpretable and steerable. Enterprise Deep Research (EDR) addresses this gap with a multi-agent framework combining adaptive planning, specialized retrieval, and human-guided reflection to enable transparent, auditable, and domain-aware research automation.

\paragraph{Multi-Agent Systems.}
The paradigm of Multi-Agent Systems (MAS)—where autonomous agents collaborate to solve problems beyond a single agent’s capability—has been revitalized by LLMs \parencite{wooldridge2009introduction}. Frameworks like AutoGen \parencite{wu2024autogen} enable flexible conversational workflows, while MetaGPT \parencite{metagpt} and ChatDev \parencite{qian2023chatdev} simulate structured team dynamics. Individual agents leverage reasoning and feedback \parencite{yao2023react}, self-reflection \parencite{shinn2023reflexion}, and external tool use \parencite{schick2023toolformer} to enhance collaboration. Yet most research focuses on open-domain or benchmark tasks. The challenge of orchestrating specialized agents for deep research across private, heterogeneous enterprise data—spanning databases, internal reports, and document repositories—remains largely unsolved.

\paragraph{Transparency and Enterprise Adoption.}
Enterprise deployment amplifies challenges of data heterogeneity, evolving knowledge, and governance, which opaque, one-shot pipelines cannot handle \parencite{choubey2025benchmarkingdeepsearchheterogeneous,lei2025spider20evaluatinglanguage}. High-stakes settings require auditable provenance, cost efficiency, and stakeholder alignment \parencite{yu2025ekrag,sambanova2025open}, making black-box agents unsuitable \parencite{aryaxai2025observability} and driving the need for interpretable systems \parencite{anthropic2024context,olah2018building}. Existing enterprise benchmarks—CRMArena-Pro \parencite{huang-etal-2025-crmarena,huang2025crmarenaproholisticassessmentllm}, OSWorld \parencite{xie2024osworldbenchmarkingmultimodalagents}, and WorkArena \parencite{workarena2024}—evaluate multi-application task execution but overlook long-horizon synthesis over proprietary data. Concurrently released, DRBench \cite{abaskohi2025drbenchrealisticbenchmarkenterprise} evaluates agents on an enterprise scale, multi-source reasoning that integrates public and private knowledge for open-ended research objectives. EDR fills this specific gap by externalizing reasoning into a transparent, modifiable plan, enabling human-in-the-loop control and auditability essential for reliable enterprise deep research.

\section{Conclusion}

We introduced Enterprise Deep Research (EDR), a multi-agent autonomous research framework that advances AI-driven enterprise analytics through steerable context engineering—a paradigm enabling dynamic and interpretable human-AI collaboration. EDR combines intelligent tool selection, adaptive planning, and cross-system retrieval to facilitate large-scale, transparent, and goal-aligned research workflows. The system achieves state-of-the-art performance on deep research benchmarks and internal enterprise evaluations, underscoring its effectiveness in complex analytical environments. This work demonstrates the potential of AI-powered automation for enterprise research and decision support, integrating advanced reasoning capabilities with enterprise-grade system design. Future work will focus on enhancing output factuality through improved citation and evidence grounding, developing predictive steering mechanisms, and expanding integration across broader enterprise data ecosystems.

\printbibliography[title={References}]

\newpage
\appendix
\section{Prompts}
\label{sec:prompts}
Here we provide all the prompts used in EDR in the order in which they are invoked during execution. 

\textbf{1. Initial task decomposition prompt. (\autoref{fig:initial_task_prompt}).   } Generates the initial \texttt{todo.md} plan by decomposing the research query into 3–5 structured, high-priority research tasks that establish the foundation for subsequent retrieval and reasoning.
\begin{figure}[h!]
\centering
\begin{tcolorbox}[
    width=\textwidth,
    left=0.5mm,
    right=0.5mm,
    top=0.5mm,
    bottom=0.5mm,
    title=Initial Task Decomposition Prompt,
    center title,
    fonttitle=\tiny\ttfamily]
\tiny
You are creating an initial research plan for the topic: "\{research\_topic\}"\\

Initial Query: "\{initial\_query\}"\\
Research Context: \{research\_context if research\_context else "Starting fresh research"\}\\

Decompose this query into 3–5 actionable research tasks. Return a JSON array with each task having:\\
• "description": Clear, actionable task (string)\\
• "priority": 1–10 (integer, higher = more important, default=5)\\
• "type": "research" (always "research")\\

Focus on: understanding the topic, gathering information, identifying key aspects, and building foundational knowledge.\\

Example for "Impacts of Generative AI on Scientific Research":\\

\begin{verbatim}
<answer>
[
  {"description": "Survey major applications of generative AI in scientific discovery", "priority": 8, "type": "research"},
  {"description": "Identify key papers and institutions leading AI-assisted science research", "priority": 7, "type": "research"},
  {"description": "Examine methodological advances enabled by generative models in ...", "priority": 6, "type": "research"},
  {"description": "Assess challenges and ethical considerations of AI-generated scientific results", "priority": 5, "type": "research"}
]
</answer>
\end{verbatim}

CRITICAL: Wrap JSON in \textless answer\textgreater tags.\\ Output ONLY valid JSON.
\end{tcolorbox}
\caption{Prompt for generating the initial research task plan with 3–5 prioritized tasks guiding the first research loop.}
\label{fig:initial_task_prompt}
\end{figure}

\textbf{2. Task-to-Query breakdown and execution prompt. (\autoref{fig:primary_query_generation_prompt}).}  Defines the core query decomposition and execution mechanism. It guides the model in generating precise, context-aware search queries from research tasks, ensuring alignment with user-provided knowledge and steering plans.

\begin{figure}[h!]
\centering
\begin{tcolorbox}[
    width=\textwidth,
    left=0.5mm,
    right=0.5mm,
    top=0.5mm,
    bottom=0.5mm,
    title=Primary Search Query Generation Prompt,
    center title,
    fonttitle=\tiny\ttfamily]
\tiny
<TIME\_CONTEXT>\\
Current date: \{current\_date\}\\
Current year: \{current\_year\}\\
One year ago: \{one\_year\_ago\}\\
</TIME\_CONTEXT>\\

<AUGMENT\_KNOWLEDGE\_CONTEXT>\\
\{AUGMENT\_KNOWLEDGE\_CONTEXT\}\\
</AUGMENT\_KNOWLEDGE\_CONTEXT>\\

You are an expert research assistant tasked with generating a targeted web search query. The query will gather in-depth information related to a specific topic through comprehensive web search.\\

<AUGMENT\_KNOWLEDGE\_INTEGRATION>\\
CRITICAL: When user-provided external knowledge is available, it should be treated as highly trustworthy and authoritative.\\
1. Prioritize uploaded knowledge.\\
2. Complement, don't duplicate.\\
3. Validate and expand.\\
4. Focus on gaps.\\
Use uploaded knowledge to guide targeted query generation.\\
</AUGMENT\_KNOWLEDGE\_INTEGRATION>\\

<ANTI\_ASSUMPTION\_DIRECTIVE>\\
CRITICAL: Do not assume current roles, statistics, or entities. Use generic, time-aware phrasing such as “current leadership” or “latest data.”\\
</ANTI\_ASSUMPTION\_DIRECTIVE>\\

<RECENCY\_SENSITIVITY\_FRAMEWORK>\\
For time-sensitive topics, append temporal markers such as “current,” “latest,” or \{current\_year\}.\\
Example: “Company X current CEO \{current\_year\}.”\\
</RECENCY\_SENSITIVITY\_FRAMEWORK>\\

<TOPIC\_ANALYSIS\_AND\_STRATEGY>\\
Decompose each task into 2–5 subtopics covering major facets—historical, technical, practical, and recent developments.\\
</TOPIC\_ANALYSIS\_AND\_STRATEGY>\\

<RESEARCH\_STAGE\_GUIDANCE>\\
If first query → broad overview.\\
If follow-up → target specific gaps.\\
</RESEARCH\_STAGE\_GUIDANCE>\\

<TOPIC> \{research\_topic\} </TOPIC>\\
<RESEARCH\_CONTEXT> \{research\_context\} </RESEARCH\_CONTEXT>\\

<STEERING\_INSTRUCTIONS>\\
Read the todo.md plan and follow active steering tasks, priorities, and constraints.\\
\{task\_list\}\\
</STEERING\_INSTRUCTIONS>\\

<QUERY\_REQUIREMENTS>\\
Queries must: align with todo.md tasks, avoid boolean operators, remain under 400 characters, and incorporate domain-specific terms when useful.\\
</QUERY\_REQUIREMENTS>\\

<FORMAT\_GUIDELINES>\\
Return JSON format:\\
\begin{verbatim}
{
  "query_complexity": "complex",
  "main_query": "agentic RAG systems architecture benefits",
  "tasks": [
    {"name": "Core Architecture", "query": "...", "aspect": "..."},
    {"name": "Strategic Retrieval", "query": "...", "aspect": "..."}
  ]
}
\end{verbatim}
</FORMAT\_GUIDELINES>\\

<AVOIDING\_ASSUMPTION\_EXAMPLES>\\
Incorrect: "John Smith Company X CEO background"\\
Correct: "current Company X CEO name {current\_year}"\\
</AVOIDING\_ASSUMPTION\_EXAMPLES>\\

<EXAMPLES>\\
Example: “streaming video platform market share 2021–2025”\\
Example: “current Country X president \{current\_year\}”\\
Example: “agentic RAG systems implementation case studies”\\
</EXAMPLES>\\

Provide your response in JSON format only.\\
\end{tcolorbox}
\caption{Prompt defining the query decomposition and execution process for targeted, time-sensitive, and context-aware research queries.}
\label{fig:primary_query_generation_prompt}
\end{figure}

\textbf{3. Reflection prompt. (\autoref{fig:reflection_prompt}). }Governs the post-synthesis evaluation stage of the research pipeline. It systematically audits coverage, identifies missing knowledge, and updates the task queue—marking completed, cancelled, or newly added search tasks. The prompt ensures iterative refinement until all critical knowledge gaps are closed.
\begin{figure}[h!]
\centering
\begin{tcolorbox}[
    width=\textwidth,
    left=0.5mm,
    right=0.5mm,
    top=0.5mm,
    bottom=0.5mm,
    title=Research Loop Reflection Prompt,
    center title,
    fonttitle=\tiny\ttfamily]
\tiny
You are an expert research evaluator analyzing a summary about \{research\_topic\}.\\

<GOAL>\\
Conduct a structured evaluation to determine:\\
1. Whether the summary is sufficiently comprehensive and accurate overall\\
2. Which SECTIONS of the summary need more detail or data\\
3. What specific knowledge gaps exist\\
4. How to formulate a targeted follow-up query to address those gaps\\
</GOAL>\\

<TODO\_DRIVEN\_REFLECTION>\\
The research loop maintains a dynamic task queue.\\
- PENDING TASKS: \{pending\_tasks\}\\
- ALREADY COMPLETED: \{completed\_tasks\}\\
- USER STEERING MESSAGES (if any): \{steering\_messages\}\\

YOUR TASK - Update the todo list:\\
1. MARK COMPLETED: Which PENDING tasks were addressed in this research loop?\\
   - Review ONLY the pending tasks listed above\\
   - Check if the current summary now covers those areas\\
   - Return their task\_ids in "mark\_completed" list\\
   - IMPORTANT: ONLY evaluate the PENDING tasks - do NOT mark tasks from "ALREADY COMPLETED"\\

2. CANCEL TASKS: Which pending tasks are no longer relevant?\\
   - Based on current findings OR user steering messages\\
   - Cancel tasks that don't align with the research direction\\
   - Return their task\_ids in "cancel\_tasks" list\\

3. ADD NEW TASKS: What critical areas still need research?\\
   - Identify knowledge gaps in the current summary\\
   - If user sent steering messages, add tasks for their requests\\
   - IMPORTANT: Check "ALREADY COMPLETED" section to avoid creating duplicate tasks\\
   - Each new task = one specific search topic\\
   - Keep tasks simple and searchable (e.g. "Research X's work at Y")\\
   - Return as objects with "description" and "rationale" in "add\_tasks" list\\
   
4. CLEAR MESSAGES: Which steering messages can be cleared?\\
   - Return list of indices of messages that have corresponding tasks created or are no longer relevant\\

RULES:\\
- ONLY mark tasks as completed if they're in the "PENDING TASKS" section above\\
- DO NOT create tasks similar to those in "ALREADY COMPLETED" section\\
- Each task = one Tavily search query\\
- Focus on WHAT to search, not HOW to organize\\
</TODO\_DRIVEN\_REFLECTION>\\

<KNOWLEDGE\_GAP\_PRIORITIZATION>\\
Use this structured framework to identify and prioritize knowledge gaps:\\

1. Gap categorization by impact type:\\
   - Critical gaps: Missing information that fundamentally undermines main conclusions\\
   - Contextual gaps: Missing background or explanatory information that would enhance understanding\\
   - Detail gaps: Missing specifics that would provide greater precision or confidence\\
   - Extension gaps: Related areas that would broaden perspective but aren't central\\

2. Prioritization matrix:\\
   - Priority 1 (Highest): Critical gaps in central topic areas directly related to the original research topic\\
   - Priority 2: Critical gaps in peripheral areas OR contextual gaps in central areas\\
   - Priority 3: Detail gaps in central areas OR contextual gaps in peripheral areas\\
   - Priority 4 (Lowest): Extension gaps OR detail gaps in peripheral areas\\
   - NEVER prioritize gaps that would lead research away from the original research topic\\
</KNOWLEDGE\_GAP\_PRIORITIZATION>\\

<QUANTITATIVE\_SUFFICIENCY\_METRICS>\\
Use these measurable criteria to objectively assess research completeness:\\

1. Coverage Completeness Score:\\
   - Assign a completion percentage to each identified subtopic:\\
     * 90-100\%: Comprehensive coverage with specific details and examples\\
     * 70-89\%: Substantial coverage but missing some specific details\\
     * 40-69\%: Basic coverage that outlines key points but lacks depth\\
     * 0-39\%: Minimal coverage or completely missing\\
   - Calculate overall coverage by averaging across all required subtopics\\
   - Research is considered "complete" ONLY when average coverage exceeds 95\% AND no critical subtopic falls below 85\%\\

2. Source Quality Assessment:\\
   - Measure proportion of information from high-authority sources\\
   - Research is considered "complete" only when at least 80\% of critical assertions are supported by Tier 1-2 sources\\

3. Evidence Density Evaluation:\\
   - For technical/scientific topics: $\geq 5$ specific metrics or measurements per key assertion\\
   - For market/business topics: $\geq 4$ quantifiable data points per major segment\\
   - For biographical topics: $\geq 5$ specific career achievements, publications, projects\\
</QUANTITATIVE\_SUFFICIENCY\_METRICS>\\

<FORMAT>\\
Return a JSON with:\\
• "research\_complete": bool (true if $\geq 95\%$ coverage, no critical gaps)\\
• "section\_gaps": map of section → missing details\\
• "priority\_section": section with most pressing gap\\
• "knowledge\_gap": concise explanation of missing info\\
• "follow\_up\_query": $\leq 400$ char search query (none if complete)\\
• "evaluation\_notes": brief overall remarks\\
• "research\_topic": unchanged original topic\\
• "todo\_updates": \{ "mark\_completed": [...], "cancel\_tasks": [...], "add\_tasks": [\{"description": "...", "rationale": "..." \}] \}\\
Task IDs must match pending list. Add new tasks only for uncovered, unique items.\\
</FORMAT>
\end{tcolorbox}
\caption{\small Structured reflection prompt for evaluating research completeness and dynamically updating \texttt{todo.md} in the Research Loop.}
\label{fig:reflection_prompt}
\end{figure}

\section{System Implementation Details}
\label{sec:system-architecture}

\subsection{API Overview}

The system exposes a comprehensive REST API built on FastAPI, providing programmatic access to all capabilities through well-designed endpoints and advanced features.

\paragraph{Research Endpoints.}
\begin{itemize}[leftmargin=1.5em, noitemsep, topsep=0pt]
    \item \texttt{POST /deep-research} – performs comprehensive research on topics with optional steering capabilities.
    \item \texttt{GET /research-status} – checks API operational status.
    \item \texttt{GET /stream/\{stream\_id\}} – provides real-time streaming updates via Server-Sent Events (SSE).
\end{itemize}

\paragraph{File Analysis Endpoints.}
\begin{itemize}[leftmargin=1.5em, noitemsep, topsep=0pt]
    \item \texttt{POST /api/files/upload} – uploads and analyzes individual files.
    \item \texttt{POST /api/files/batch-upload} – uploads multiple files simultaneously.
    \item \texttt{GET /api/files/\{file\_id\}/analysis} – retrieves analysis results.
    \item \texttt{POST /api/files/\{file\_id\}/analyze} – manually triggers analysis.
    \item \texttt{GET /api/files/\{file\_id\}/content} – downloads original files.
    \item \texttt{GET /api/files/\{file\_id\}/status} – checks processing status.
    \item \texttt{GET /api/files/} – lists all uploaded files with filtering options.
    \item \texttt{DELETE /api/files/\{file\_id\}} – removes files from the system.
  \end{itemize}

\paragraph{Database Integration Endpoints.}
\begin{itemize}[leftmargin=1.5em, noitemsep, topsep=0pt]
    \item \texttt{POST /api/database/upload} – uploads SQLite, CSV, and JSON database files.
    \item \texttt{GET /api/database/list} – retrieves all uploaded databases.
    \item \texttt{GET /api/database/\{database\_id\}/schema} – accesses database schema information.
    \item \texttt{POST /api/database/query} – executes natural language queries against databases.
    \item \texttt{DELETE /api/database/\{database\_id\}} – removes uploaded databases.
  \end{itemize}

\paragraph{Interactive Steering Endpoints.}
\begin{itemize}[leftmargin=1.5em, noitemsep, topsep=0pt]
    \item \texttt{POST /steering/message} – sends natural language steering messages during research.
    \item \texttt{GET /steering/plan/\{session\_id\}} – retrieves current research plans in Markdown format.
    \item \texttt{GET /steering/status/\{session\_id\}} – monitors real-time plan status.
    \item \texttt{GET /steering/interactive/session/\{session\_id\}} – supports frontend polling compatibility.
    \item \texttt{GET /steering/sessions} – lists all steerable research sessions.
    \item \texttt{GET /steering/examples} – accesses natural language steering examples.
  \end{itemize}

Advanced API features include streaming responses using Server-Sent Events (SSE) with heartbeat mechanisms for real-time updates, background task processing with asynchronous execution and status tracking, comprehensive error handling with detailed error responses and recovery suggestions, CORS support for cross-origin resource sharing in web applications, automatic API documentation generation using OpenAPI/Swagger standards accessible at /docs, cache control middleware for optimal performance, React frontend integration with static file serving and catch-all routing, multiple LLM provider authentication methods including OpenAI, Anthropic, Groq, and SambaNova API keys configured through environment variables, and rate limiting with configurable delays and exponential backoff for API protection.

\subsection{Frontend Architecture}
The frontend architecture is implemented using React 18 and TypeScript, leveraging concurrent rendering, automatic batching, and improved error boundaries for high performance and robust developer experience. TypeScript ensures comprehensive type safety, strict type checking, and seamless refactoring. The interface design employs Material-UI for accessible, responsive, and themable components that adhere to WCAG guidelines, complemented by advanced UI elements such as sortable data tables, validated forms, and interactive visualizations. Tailwind CSS provides a utility-first styling framework with custom design tokens, responsive breakpoints, and optimized CSS generation for minimal bundle sizes, along with dark mode and accessibility utilities. Data fetching and state synchronization are managed by React Query, which supports intelligent caching, background refetching, optimistic updates, and automatic retry mechanisms to ensure a smooth and responsive user experience.

\end{document}